\documentclass{article}

\usepackage[nonatbib, final]{neurips_2021_ml4ad}

\usepackage[utf8]{inputenc} 
\usepackage[T1]{fontenc}    
\usepackage{hyperref}       
\usepackage{url}            
\usepackage{booktabs}       
\usepackage{amsfonts}       
\usepackage{nicefrac}       
\usepackage{microtype}      
\usepackage{xcolor}         
\usepackage{epsfig}
\usepackage{graphicx}
\usepackage{multirow}
\usepackage{amsmath}
\usepackage{amssymb}
\usepackage{gensymb}
\usepackage[numbers,sort,compress]{natbib}
\usepackage{siunitx}
\usepackage{caption}
\usepackage{subcaption}
\usepackage{authblk}

\newcommand{\figref}[2][{}]{\hyperref[#2]{\figurename~\ref{#2}#1}} 
\newcommand{\tableref}[2][{}]{\hyperref[#2]{\tablename~\ref{#2}#1}} 

\title{Scalable Primitives for Generalized Sensor Fusion in Autonomous Vehicles}

\author{\textbf{Sammy Sidhu}}
\author{\textbf{Linda Wang}}
\author{\textbf{Tayyab Naseer}}
\author{\textbf{Ashish Malhotra}}
\author{\textbf{Jay Chia}}
\author{\textbf{Aayush Ahuja}}
\author{\\\textbf{Ella Rasmussen}}
\author{\textbf{Qiangui Huang}}
\author{\textbf{Ray Gao}}
\affil{Woven Planet Level 5}

\begin{document}

\maketitle
\begin{abstract}
In autonomous driving, there has been an explosion in the use of deep neural networks for perception, prediction and planning tasks. As autonomous vehicles (AVs) move closer to production, multi-modal sensor inputs and heterogeneous vehicle fleets with different sets of sensor platforms are becoming increasingly common in the industry. However, neural network architectures typically target specific sensor platforms and are not robust to changes in input, making the problem of scaling and model deployment particularly difficult. Furthermore, most players still treat the problem of optimizing software and hardware as entirely independent problems. We propose a new end to end architecture, Generalized Sensor Fusion (GSF), which is designed in such a way that both sensor inputs and target tasks are modular and modifiable. This enables AV system designers to easily experiment with different sensor configurations and methods and opens up the ability to deploy on heterogeneous fleets using the same models that are shared across a large engineering organization. Using this system, we report experimental results where we demonstrate near-parity of an expensive high-density (HD) LiDAR sensor with a cheap low-density (LD) LiDAR plus camera setup in the 3D object detection task. This paves the way for the industry to jointly design hardware and software architectures as well as large fleets with heterogeneous configurations.

\end{abstract}

\vspace{-.15in}
\section{Introduction}
\vspace{-.1in}
In a very short amount of time, Deep Neural Networks (DNNs) have become the dominant approach for solving various problems in the computer vision domain~\cite{lecun2015deeplearning, alexnet}. This is especially true for the Autonomous Vehicle (AV) space where we see that the winning approach for nearly every benchmark dataset is Deep Learning based~\cite{kitti, waymo_open_dataset, lyft2019}. These datasets are composed of a variety of sensors such as LiDAR, RaDAR, and cameras and target a variety of tasks required for autonomous driving such as object detection~\cite{kitti, waymo_open_dataset, lyft2019}, object tracking~\cite{nuscenes2019}, semantic segmentation~\cite{nuscenes2019} and motion prediction~\cite{lyft2020prediction, waymo_open_dataset}. As a product of these public datasets, we have seen an explosion of DNN model development in literature (discussed in section \ref{related_work}), where we are seeing the bar being raised year over year on these tasks. However we often see that these cutting edge models usually only operate on one modality and one task such as in \cite{imvoxelnet, lift-splat} or that they perform multi-modal sensor fusion in a way that the sensor inputs are entangled and non-modular\cite{Chen2017Multiview3O, Liang2018DeepCF, mvx-net, qi2017frustum, dscnet}.

For many AV companies, the ability to scale a unified model architecture across sensor configurations, platforms, and engineering teams can outweigh the benefits of deploying an over-optimized one for a single specific platform. Ideally, we want our model architecture to be designed in a way where both sensor inputs and target tasks are modular and easily switchable.  This many-to-many relationship allows models to be trained with one or multiple inputs such as images, LiDAR, or RaDAR to produce one or more of the aforementioned tasks. To achieve this, we introduce a Generalized Sensor Fusion (GSF) meta-architecture in section \ref{gsfsection} that we use for all our experiments. Using the GSF meta-architecture, we demonstrate 3D object detection results leveraging various methods in the form of encoders such as a LiDAR PointNet voxelizer, Stereo Cost Volume and a multiview image uplifting method that we discuss in section \ref{raycastsection}.

Using the methods we implement using the GSF framework, we run ablations with various sensor configurations both for front facing and surround detection. With these results, we draw several important insights on how to design better sensor configurations and created a meta-architecture with powerful abstractions that is used by a large engineering team and can be deployed on heterogeneous sensor platforms.
\vspace{-.1in}
\section{Related Work}
\vspace{-.1in}

\label{related_work}
3D object detection using multiple sensor configurations has gained tremendous attention in recent years. We categorize relevant research in the following areas and build upon some of the working principles proposed in these approaches.
\vspace{-0.1in}
\subsection*{Monocular and Stereo Detection}
\vspace{-0.1in}

Monocular and Stereo 3D detection methods can be broadly categorized into three domains: Direct, Voxel grid-based and Pseudo-LiDAR based.

\textit{Direct}: Direct methods directly estimate 3D bounding boxes from images without generating an intermediate 3D feature representation. As there is no explicit depth-based localization of features in 3D space, these approaches geometrically reason for 3D objects by associating object keypoints in image plane. M3D-RPN\cite{m3drpn} proposed to use depth-aware convolutions using non-shared kernels for each sub region. This enables the approach to learn location specific features that are correlated in 3D space. Stereo R-CNN~\cite{stereocnn} formulates 3D detection into multiple components explicitly to resolve several constraints like depth and matching keypoints. This approach suffers from generalization to other agent types and occlusions in image space.

\textit{Voxel Grid-based}:
Grid-based methods generally voxelize the world observed by the cameras into 3D grid cells and populate these cells with image features. These features are then further processed to generate 2D BEV features for efficient downstream processing. Such methods provide rich geometric reasoning to estimate 3D objects. Monocular methods like~\cite{oftnet, mono2dto3dliftinig} project image features to 3D voxels to reason downstream 3D object detection task in 3D space. Such monocular methods lack the ability of localizing features in a frustum due to depth ambiguity. CaDDN~\cite{reading2021categorical} copes with this problem by estimating depth distribution for each pixel before projecting the features to 3D. Similarly for stereo, DSGN~\cite{chen2020dsgn} builds a plane sweep volume (PSV) distributed using stereo image features to estimate per pixel depth as an auxiliary task in addition to 3D object detection. It warps image-centric PSV features to 3D geometric space to perform 3D object detection in a end-to-end manner. PLUME~\cite{wang2021plume} avoids building the PSV and instead estimates occupancy of 3D voxel grid as the auxiliary task. This makes the model memory and compute efficient as it does not build an explicit 3D cost volume for depth estimation. Our stereo methods in section \ref{stereo_experiments} draws inspiration from both DSGN and PLUME.

\textit{Pseudo-LiDAR based}:
The pioneer work in this area ~\cite{wang2019pseudo,pseudolidar++} generates depth information from single or stereo images and leverage existing 3D LiDAR detectors for 3D object detection. These approaches are not learned in end-to-end manner and train both the depth estimation and object detection branches separately. Such approaches suffer from learning intermediate features and depth maps that should be rather object centric than pixel centric.~\cite{ple2e} transforms this approach to a end-to-end trainable model, so that both detection and depth losses are optimized jointly. This shows better performance than disjoint Pseudo-LiDAR methods. Ma~\cite{ma2019accurate} combines RGB features with the Pseudo-LiDAR pipeline by introducing a particular component that maps 2D image data to 3D point cloud.

\vspace{-0.1in}
\subsection*{Multiview Monocular Detection}
\vspace{-0.1in}

Multiview monocular-based methods take multiple image inputs that can be stitched together to form a birds-eye-view (BEV) representation of the scene around the ego. Multiview 3D detection methods first extract features from the images, then lifts the features to a 3-dimensional frame, most often a voxel representation, that can be shared across all cameras. The challenge of projecting from 2D to 3D is that depth is required, however, the depth associated with each pixel is ambiguous. To overcome this challenge, \cite{lift-splat} generates possible depth distributions for each pixel, which makes it able to place the context in the nearest pillar, rather than projecting along the entire ray. \cite{imvoxelnet} projects the 2D image features into an aggregated 3D voxel representation using a pinhole camera model. The 3D representations are then processed by a standard CNN to predict BEV bounding boxes.

\vspace{-0.1in}
\subsection*{LiDAR and Vision Fusion Detection}
\vspace{-0.1in}

Multimodal fusion between LiDAR point clouds and images aims to combine the strengths of each modality. However, an end-to-end trainable model with both camera and LiDAR sensor modalities is a non-trivial task since images represent a dense 2D projection of the scene, while LiDAR captures a sparse 3D structure. For many methods, feature maps are first learned separately for each modality \cite{Chen2017Multiview3O, Ku2018Joint3P, mvx-net, Liang2018DeepCF}. After features are learned from LiDAR and images, \cite{Chen2017Multiview3O} then uses the BEV map to generate 3D object proposals to fuse image and point cloud features through ROI-pooling. Similarly, \cite{Ku2018Joint3P} uses a regional proposal network on BEV and image feature maps to generate region proposals for each modality. The proposals are then passed to a detection network to fuse the modalities. Both of these approaches perform late fusion, which inhibits the network from learning interactions of the two modalities at earlier stages. To learn the interactions at earlier stages, \cite{mvx-net} fuses LiDAR points and image features by appending each 3D point with the corresponding image feature. In order to reduce the number of image features dropped, \cite{mvx-net} appends the image features after the point cloud has been voxelized. \cite{Liang2018DeepCF} uses a continuous fusion layer in order to bridge multiple intermediate feature layers and perform multi-sensor fusion at different scales. However, these architectures are LiDAR dependent and are not easily interchangeable between modalities.

\vspace{-.1in}
\section{Generalized Sensor Fusion}
\vspace{-0.1in}
\label{gsfsection}
\begin{figure*}[th!]
\begin{center}
   \includegraphics[width=0.8\linewidth]{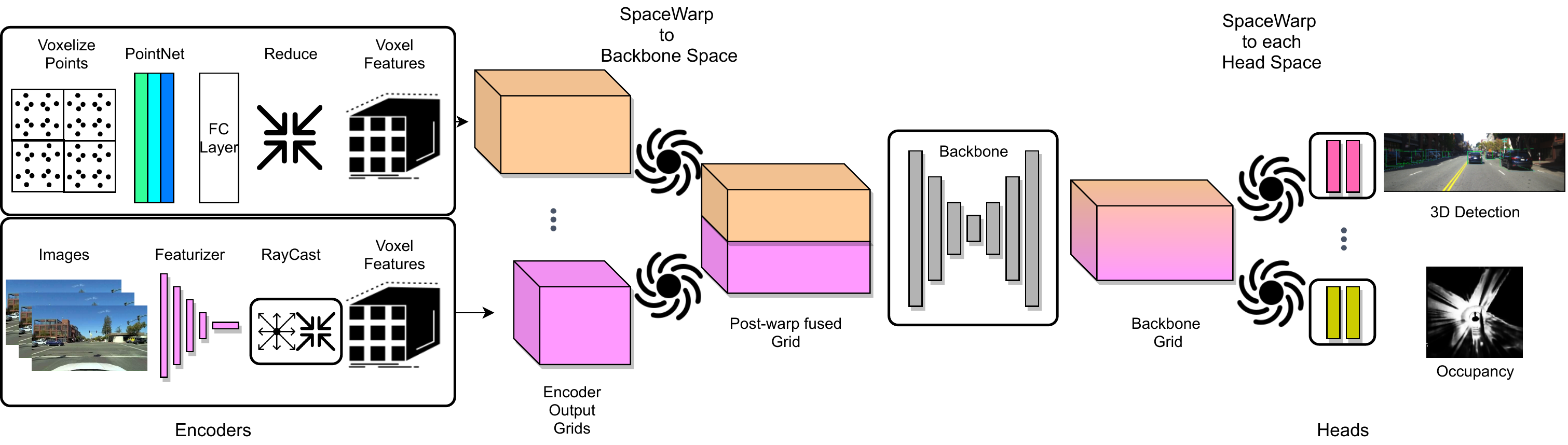}
\end{center}
   \caption{Overview of the Generalized Sensor Fusion (GSF) meta-architecture that allows for a modular set of both sensors and tasks.}
\label{fig:metaarch}
\end{figure*}
To experiment with various sensors, methods and tasks in a modular fashion, we introduce a set of primitives called Generalized Sensor Fusion (GSF) that allow for a scalable meta-architecture. An overview can be seen in \figref{fig:metaarch}.
\begin{itemize}
    \vspace{-0.05in}

	\item a {\bf Space} defines the coordinate system of an abstract space. An example of this is a Cartesian Space that is defined by voxel range and resolution.
	\vspace{-0.05in}
	\item a {\bf Pose} defines the origin and coordinate basis change relative to a base pose. This is useful when dealing with multiple sensors in different locations in the same time-step or dealing with sensors across time-steps accounting for ego motion.
	\vspace{-0.05in}
	\item a {\bf Grid} is an object that abstracts a 5 dimensional feature tensor $(N, C, Z, X, Y)$ and  attaches it to both a {\bf Space} and {\bf Pose} that defines its coordinate system.
	\vspace{-0.05in}
	\item {\bf SpaceWarp} allows for the transformation of a source {\bf Grid} into another {\bf Space} and {\bf Pose}, yielding a new Grid in the target Space and Pose. This allows for parts of the network to operate in different Spaces and the ability to convert one to another in an efficient and differentiable way. SpaceWarp operates by first querying the target Space object for 3D Grid sample points that we would need to transform the source grid. Then we correct for Pose and coordinate basis changes, which can also be chained if there are multiple intermediate Spaces between the source and target Space. Finally, we perform a 3D tri-linear interpolation using the corrected sample points to warp the source Grid to the target Space.
	\vspace{-0.05in}
	\item {\bf Encoders} are responsible for transforming source sensor inputs, such as camera images or LiDAR point clouds, into featurized {\bf Grids}. An example of converting camera images into a Cartesian Grid is described in section \ref{raycastsection}. For a LiDAR point cloud input, an example encoder may be a PointNet sub-network~\cite{voxelnet} that yields a Grid composed of PointNet features for every 3D voxel. One thing to note is that a model may have multiple encoders each running in their own Space and Pose. Additionally, Encoders may also produce auxiliary outputs and losses that are used in downstream parts of the network. 
	\vspace{-0.05in}
	\item {\bf Heads} are responsible for taking in a {\bf Grid} and producing outputs for a Task such as object detection or segmentation. Head also can be parameterized by the {\bf Space} they operate in and will {\bf SpaceWarp} the input Grid to the target space as needed. This allows various Heads in the network to produce predictions in their own Space without modification to the {\bf Encoders} or {\bf Backbone}. An example is if the detection and segmentation tasks operate in different voxel cell sizes compared with each other and the Backbone. 
	\vspace{-0.05in}
	\item The {\bf Backbone} is a shared sub-network that ingests all {\bf Encoder} outputs and feeds all the downstream {\bf Heads}. The input for the Backbone is a sequence of {\bf Grids} that each get {\bf SpaceWarped} to common {\bf Space} defined by the Backbone. Then these warped Grids are fused and fed into a sub-network that is user defined. Finally the Backbone yields a Grid that may be in a different Space that is fed to each of the heads.
\end{itemize}
\vspace{-0.1in}
The net result allows us to implement encoders and tasks in a self contained manner. For encoders,  we show a PointNet and a vision uplifting example in \figref{fig:metaarch} and for tasks, we show a detection and occupancy task example. 

\vspace{-.1in}
\section{Raycasting Vision uplift method}
\vspace{-.1in}

\begin{figure*}[th!]
\begin{center}
   \includegraphics[width=0.75\linewidth]{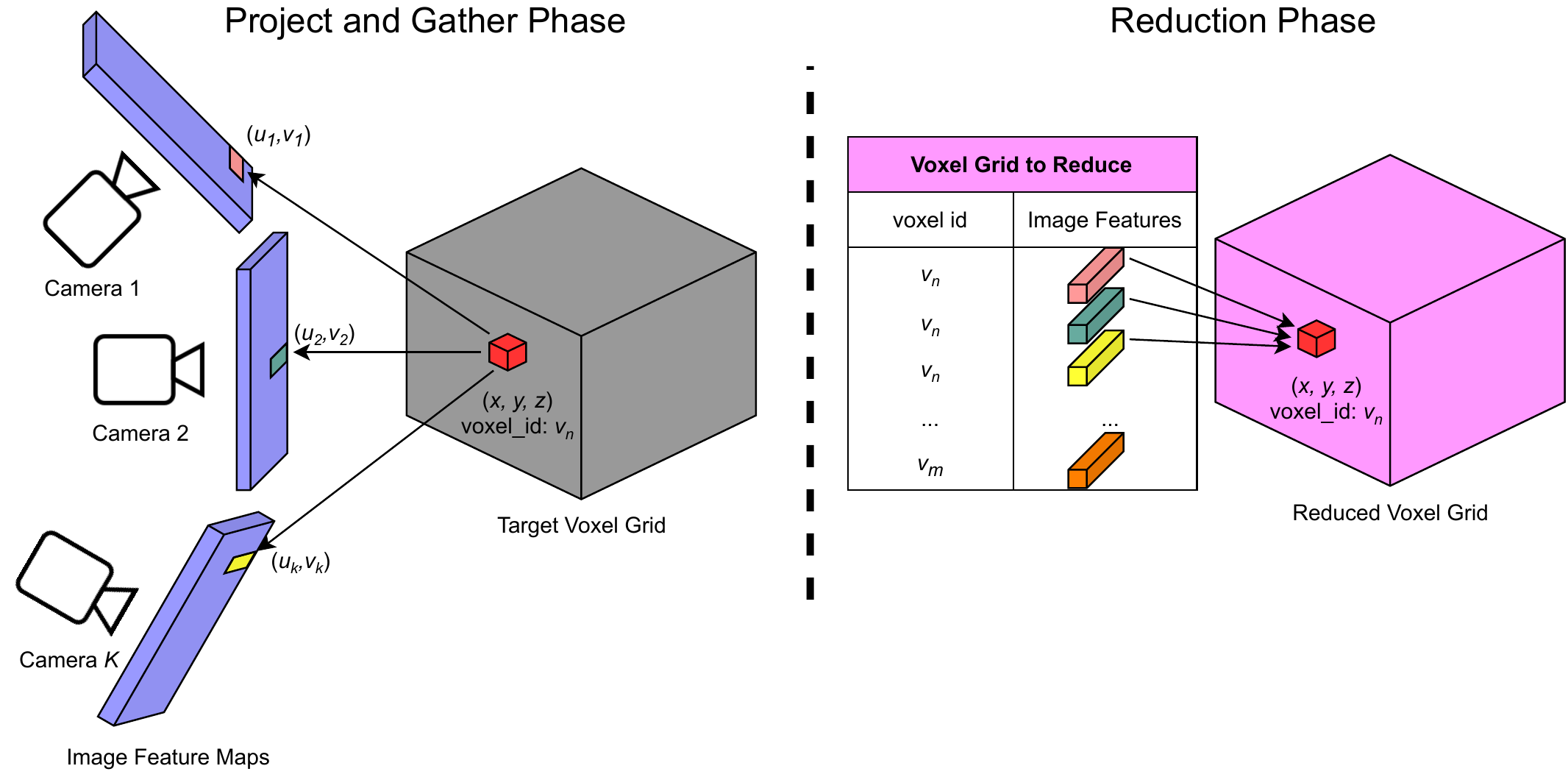}
\end{center}
   \caption{Overview of the Raycasting operation that maps a set of image features to a 3D grid.}
\label{fig:raycast}
\end{figure*}

\label{raycastsection}
Image features are inherently 2D, however, in section \ref{gsfsection}, all sensor encoders are required to have the same output space. To uplift the image features to a 3D Cartesian Space, we use a method called Raycasting. The Raycasting operation is parameterized by the target range and voxel resolution and during computation will take in a set of image feature maps from the same frame as well as their corresponding metadata such as pose and the projection matrix. The Raycast operation is comprised of two phases, the first is the project and gather phase and the second is the reduction phase. A diagram of this method is shown in \figref{fig:raycast}.

In the project and gather phase, we first project the centroid of each of the target voxels onto each of the inputted cameras that has that voxel in its field of view. The result of this projection is a tensor that stores the $(voxel\ id, camera\ index, u, v)$ of each valid camera voxel pairs. The second and final step of this phase is to gather the features from the image sub-network for each of the matched valid camera voxel pairs using the previously computed image coordinates; this produces a tensor that is size $(valid\ voxel\ pairs, image\ feature\ channels)$. We also store the corresponding $voxel\ id$ for each of these pairs for the reduction phase. Both of these operations are vectorized and are fast to compute on GPU without having to rely on any custom kernels. Another thing to note here is that we can have anywhere from zero to the number of cameras occurrence $K$ for a $voxel\ id$. A $voxel\ id$ will be absent from our feature tensor in the case that it doesn't fall on any of the inputted cameras, and we may have $K$ occurrences in the case that it falls on every camera.

In the reduction phase, we use the gathered feature tensor and corresponding $voxel\ id$s and perform a scatter reduce, which results in a tensor that is size $(num\ voxels, image\ feature\ channels)$ and has the accumulated feature vectors. For absent $voxel\ id$s, the result of the scatter for that row will be a zero vector. Next the reduced voxels are reshaped to transform it into a 3D grid. Finally we concatenate the voxel location in Cartesian coordinates as additional channels to the final grid. This ensures voxels that only fall on a single camera and share a frustum with another voxel don't produce the same result, which could be harmful for a task like object detection.

This method for uplifting differs from the one presented in~\cite{lift-splat} due to the fact that we do not encode our image features as a depth volume prior to projection and that we project the voxels onto the image plane rather than shooting out the image space depth cells. This actually yields a much higher density in the voxel grid since every voxel that is in a field of view of a camera will have image features. This method also differs from~\cite{wang2021plume} since Raycast runs in a sparse fashion and allows for a generic camera setup with or without camera overlap rather than being constrained to stereo. 

\vspace{-.1in}
\section{Datasets}
\vspace{-.1in}

\label{datasets}
In order to evaluate the performance of our proposed method and perform ablations on sensors, we benchmark it on 2 datasets:
\begin{itemize}
 \vspace{-0.05in}
 \item the Panoramic with LiDAR dataset: each data sample consists of a LiDAR point cloud and multiple monocular cameras covering 360$^{\circ}$  of the surroundings.
 \vspace{-0.05in}
 \item the Forward Stereo with LiDAR Dataset (FSLD): each data sample consists of a LiDAR point cloud and a front-facing stereo camera image pair.
  \vspace{-0.1in}
\end{itemize}

\begin{figure}
\centering
\begin{minipage}[t]{.45\textwidth}
  \centering
  \includegraphics[width=1.0\linewidth]{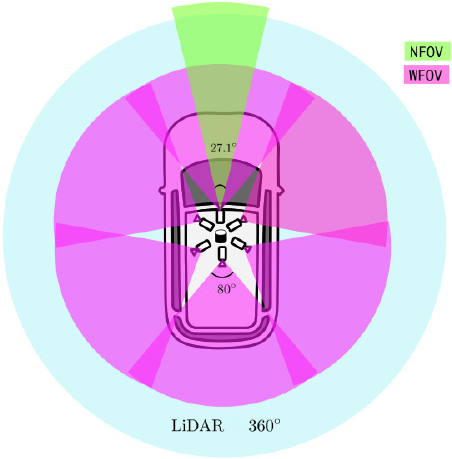}
  \captionof{figure}{This figure illustrates the AV setup for the dataset described in section \ref{panoramicdataset}, where we have 6 wide field of view (WFOV) cameras and 1 narrow (NFOV) in addition to surround LiDAR.}
  \label{fig:multiview_setup}
\end{minipage}%
\hfill
\begin{minipage}[t]{.45\textwidth}
  \centering
  \includegraphics[width=1.0\linewidth]{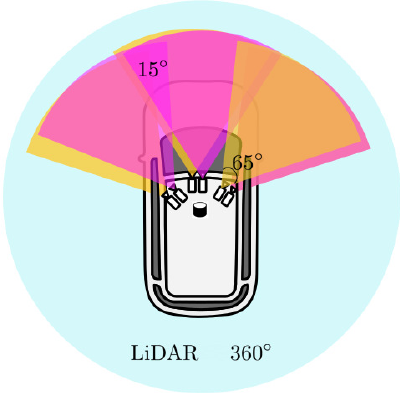}
  \captionof{figure}{This figure illustrates our multi-stereo setup with 3 pairs. However we focus only on the forward facing pair.}
  \label{fig:stereo_setup}
\end{minipage}
\end{figure}

\vspace{-0.1in}
\subsection{Panoramic with LiDAR Dataset}
\vspace{-0.1in}

\label{panoramicdataset}

The dataset consists of 41,525 training samples and 8,998 test samples, with a total of 860,075 car annotations. The data is collected in areas spanning across San Francisco Bay Area. Each data sample includes a LiDAR point cloud and 7 monocular cameras covering 360$^{\circ}$  of the surrounding. The cameras have a resolution of 1080$\times$1920 pixels but we feed the models using this dataset half-res images. \figref{fig:multiview_setup} presents the details of the sensor layout.

To assess the model performance on LiDAR, our experiments compare high and low density LiDAR configurations, which are defined below. The low density configuration is similar to LiDARs we might see on high volume consumer vehicle for ADAS soon. The low density configuration is generated by sub-sampling data from the full density panoramic dataset LiDAR via vertical beam sub-sampling, azimuth sub-sampling and probability of detection mapping.

\begin{itemize}
	\vspace{-0.1in}
    \item High density (HD): Single roof LiDAR with 64 beams.
    \item Low density (LD): Single roof LiDAR with 13 beams and azimuth subsampling of 10.
    \vspace{-0.1in}
\end{itemize}

The following table details specifications for each sensor configuration, including angular resolution, number of LiDAR beams, and maximum number of points per frame. 
\begin{table}[h!]
\small
\begin{center}
 \begin{tabular}{|c | c | c| c| c|} 
 \hline
 Config 
 & \multicolumn{1}{|p{1cm}|}{\centering angular step (\degree)}
 & \multicolumn{1}{|p{1cm}|}{\centering number beams}
 & \multicolumn{1}{|p{1.75cm}|}{\centering max points/scan}
 & \multicolumn{1}{|p{1.75cm}|}{\centering median points/object}\\ [0.5ex] 
 \hline\hline
 HD & 0.2 & 64 & 113,400 & 117 \\
 \hline
 LD & 2 & 13 & 2,340 & 1 \\
 \hline
\end{tabular}
\end{center}
\caption{LiDAR Dataset Configurations. Median points per object is median number of points that are contained in annotated objects.
\label{lidardataset}}
\end{table}

\vspace{-.1in}
\subsubsection*{Valid Objects}
\vspace{-.1in}

Our training data contains annotations for objects across occlusion as well. As such, many objects are not visible in the camera frame. This occlusion could occur from all sorts of objects such as other cars or static obstacles such as bushes. To prevent penalizing the model during training and test for not detecting these objects that are not visible, we derive a set of valid objects using occlusion reasoning. This is done by constructing a range image of the LiDAR point cloud and deriving the percentage of the annotation box visible from the camera frame. Only boxes above a certain threshold are actually used. For objects below the visibility threshold, they do not contribute any loss during training and are ignored during metric calculation.

\vspace{-.1in}
\subsection{Forward Stereo with LiDAR Dataset}
\vspace{-.1in}

\label{stereodatasets}
The goal of the Forward Stereo with LiDAR (FSLD) dataset is to evaluate the impact of multi-view stereo. The sensor layout, shown in \figref{fig:stereo_setup} is similar to the panoramic layout except that there are 3 forward/side stereo cameras pairs rather than 7 monocular cameras.
Although, we have 3 camera pairs, we only focus on the forward facing one for our ablations.
 Front stereo consists 25000 training images and 3750 test frames. Each image has a 65$^{\circ}$ horizontal field of view and a resolution of 720$\times$1536 pixels. A single stereo camera has a baseline of 35.6~\si{cm} as compared to 54~\si{cm} in KITTI~\cite{kitti} sensor setup. Although wider baseline cameras enable higher quality depth estimations at far ranges, these are mechanically less viable due to physical constraints in multi-view stereo setups on the cars.

\vspace{-.1in}
\section{Panoramic Experiments and Results}
\vspace{-.1in}

\label{panoramic_experiments}
Using our modular GSF meta-architecture, we evaluate detection performance on a variety of configurations of vision and LiDAR. For models with LiDAR, we use a 2 layer PointNet encoder, as shown in \figref{fig:metaarch}, that creates a learnable embedding for each voxel. All LiDAR models use the same encoder in the range $(X, Y, Z)$ of $[-51.2m, -51.2m, -2m]$ to $[+51.2m, +51.m, 12m]$ with a voxel cell size of $[0.32m, 0.32m, 14m]$. 

For models with vision, we use our Raycast method that is fed image features from a ResNet-18~\cite{resnet} hour-glass model. The Raycaster then uplifts the image features to the same range as the LiDAR encoder but at voxel cell size of $[0.4m, 0.4m, 1m]$, yielding voxels in 3D rather than pillars. 

For all models, we use a 3 stage ResNet style hour-glass shared backbone that SpaceWarps each of the encoder outputs to the common space which is the same as the LiDAR encoder space. All models also use the AFDet~\cite{afdet} head for 3D object detection. We do not use any augmentations during train or test time for these experiments. This is an area for future improvements.
\vspace{-0.1in}
\subsection*{Sensor Configuration Ablation}
\vspace{-0.1in}
The three LiDAR profiles that we sweep over are High Density (HD), Low Density (LD), and None. In addition, we evaluate the effects of using vision in combination with LiDAR. The vision models here take in 6 wide field of view (WFOV) camera images for each frame providing full $360^\circ$ panoramic coverage as seen in \figref{fig:multiview_setup}. We also determine if having LiDAR as supervision during training only in the form of an occupancy grid task improves performance for vision only models at test time.

\begin{figure}
\centering
\begin{minipage}{.45\textwidth}
\begin{tabular}{|c|c|c|c|}
\hline
LiDAR & Vision  & AP@0.5 & Latency \\ \hline
\multirow{2}{*}{HD}       &                            & 93.4   &   16.6                \\ \cline{2-4}
                                         & X                          & \textbf{98.3}   &   49.9                 \\ \hline \hline
\multirow{2}{*}{LD}        &                            & 64.6      &   \textbf{16.0}                \\ \cline{2-4}
                                         & X                          & 91.3   &   49.8                \\ \hline \hline
None                                     & X                          & 65.0   &   51.0                \\ \hline
HD*                         & X                           & 66.9   &   53.5                \\ \hline
\end{tabular}
\caption{Results for sensor configuration ablations. We report the Bird's Eye View (BEV) Average Precision at 0.5 IoU as well as the inference latency in milliseconds on a NVIDIA V100. All of these models are trained/evaluated the same datasets but have different simulated configurations as described in \ref{panoramicdataset}. \textit{HD*} signifies LiDAR was only used during train time.
\label{Tab:multiview_results}}
\end{minipage}%
\hfill
\begin{minipage}{.5\textwidth}
  \centering
  \includegraphics[width=1.0\linewidth]{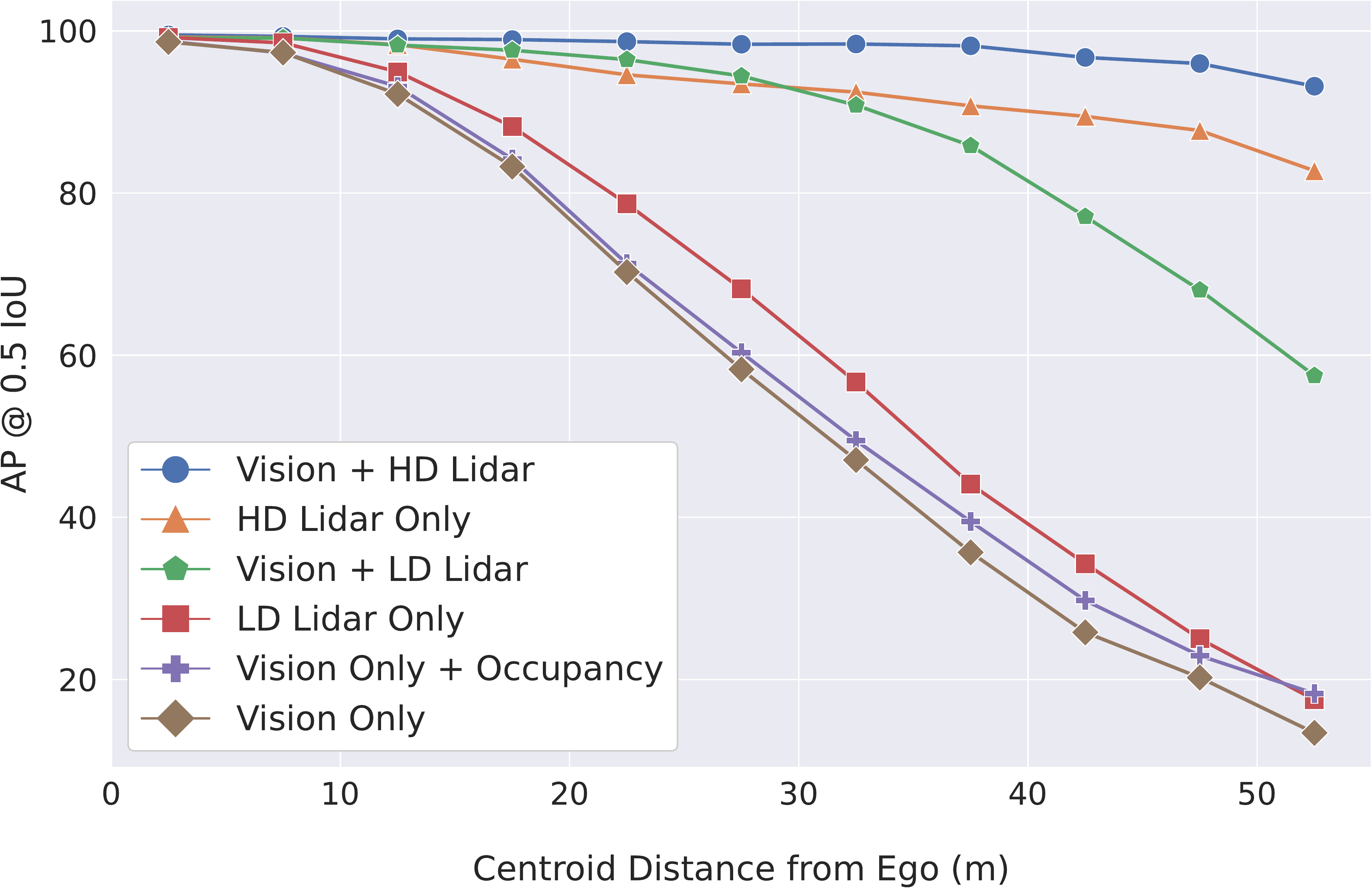}
  \captionof{figure}{AP vs Distance: We measure Car AP@0.5 BEV IoU (for non-heavily-occluded objects) as a function of distance of objects from the ego in 5m radial buckets.}
  \label{fig:vl_distance}
\end{minipage}
\end{figure}

In \tableref{Tab:multiview_results}, we can view the results of our various experiments with LiDAR and Vision. We establish baselines for single sensor models for LD LiDAR, HD LiDAR and panoramic vision. As expected, we see that with higher density LiDAR, our AP for LiDAR only models improves massively (+28 AP). We also see that the inclusion of vision to the model can substantially improve performance (+26AP for LD) but becomes less significant for the \textit{HD LiDAR} model (+5AP). Next, we observe that the addition of LD LiDAR to the vision only model can improve performance significantly (+26AP). Finally, we can conclude that single modality models perform much worse compared to their multi-modal siblings. We also see that the \textit{LD + vision} model performs nearly at the level of a model that only uses an expensive HD LiDAR at a fraction of the cost. 
\vspace{-0.1in}
\subsection*{Multiview}
\vspace{-0.1in}
\label{multiview_experiments}
In this experiment, we evaluate the effects of overlapping cameras by comparing the performance of the Raycast method on Multi-View (non-stereo) cameras vs. a monocular camera. In the panoramic dataset, we have input images from both WFOV cameras (FOV: 80$^{\circ}$, FL: 3mm) and a NFOV camera (FOV: 27$^{\circ}$, FL: 12mm). Somethings to note are that the baseline between the forward facing cameras is small and the NFOV camera frustum is completely contained in the WFOV camera (see \figref{fig:multiview_setup}).

\begin{table}[t]
\begin{center}
 \begin{tabular}{|c | c | c|} 
 \hline
 Run & AP @ IoU=0.5 & Mean DE (cm) \\ [0.5ex] 
 \hline\hline
 Mono & 65.0 & 62 \\ 
 \hline
 Multiview (MV) & 69.0 & 54 \\
 \hline
\end{tabular}
\vspace{0.1in}
\caption{Mono Raycasting (Mono) vs. MultiView Raycasting (MV). The metrics are computed only on objects in the overlapping region of the both camera frustums within a radius of 50m from the ego vehicle. Both AP and Mean DE (Mean depth error) see improvements in the MV case. Mean DE is absolute mean of the ground truth and prediction depth of matched objects only.
\label{Tab:overlapping_cams}}
\end{center}

\end{table}

Since the NFOV camera covers less area, the results presented in \tableref{Tab:overlapping_cams} are filtered to objects that are visible to both cameras. Based on the results, multiview shows an improvement in both AP (+4AP) and localization errors (-8~cm mean DE) of matched true positive objects. 
\vspace{-0.15in}
\subsection*{Occupancy Prediction}
\vspace{-0.1in}
\label{sect:occupancy_pred}
In addition to evaluating overlapping cameras, we also evaluate the performance gain from including an auxiliary occupancy prediction task during model training for vision only. Recent literature in multitask learning has demonstrated the ability of leveraging information learned from one task to another. By properly balancing losses from each task, research has shown that multitask models can outperform separately trained models~\cite{Kendall2018MultitaskLU, Chen2018GradNormGN, Liu2019EndToEndML}.

The auxiliary task produces a 3D voxel grid, where each voxel's value represents the probability of the voxel being occupied. Ground truth for this task is obtained by voxelizing the point cloud obtained from our high-density LiDAR data, and labelling all voxels with at least one point as occupied. The task is performed by the occupancy prediction head, which consists of two 2D convolution layers. The last layer outputs $Z$ channels that represent the probability of that voxel being occupied. We observe improvements to detection performance at further ranges as seen in \figref{fig:vl_distance}.

\vspace{-0.1in}
\subsection*{Training Details}
\vspace{-0.1in}

All models in this section are trained for 60 epochs on the panoramic dataset using the Adam optimizer~\cite{Adam} and the OneCycle Learning Rate policy. We use 64 NVIDIA V100 16GB to train each model with a batch size of 6 frames or 36 images per GPU.

\vspace{-0.1in}
\section{Stereo Experiments and Results}
\vspace{-0.1in}
\label{stereo_experiments}

For our experiments on the stereo dataset, we evaluate 3 primary approaches as GSF encoders. All of the approaches share the same ResNet style image featurizer, backbone and detection head but differ in how the image is uplifted.   
\begin{itemize}
	\vspace{-0.05in}
    \item \textbf{Image-Centric stereo Cost Volume (CV)}:
For every stereo pair, we correlate the left and warped right features to construct 3D camera frustum volume, similar to the plane sweep volume in~\cite{chen2020dsgn}. We then warp this cost volume to a Cartesian Grid that runs in the same GSF pipeline as the models in prior section. We also evaluate two methods to add an auxiliary loss for depth estimation. The first is a direct per-pixel depth estimation that uses LiDAR points projected onto the image as supervision. The second is an unsupervised task that minimizes the image reconstruction error similar to the work in~\cite{monodepth2}.
	\vspace{-0.05in}
    \item \textbf{Raycast multiview with stereo cameras}:
Next we use the Raycast method that we discussed in section \ref{raycastsection}, but use it on the largely overlapping forward facing stereo cameras. Additionally, we evaluate the use of the occupancy auxiliary task that was discussed in \ref{sect:occupancy_pred}.
	\vspace{-0.05in}
    \item \textbf{Raycast multiview with a monocular camera}:
Finally, try the same approach as above but provide the Raycaster a single image. We achieve this by simply dropping the right image of our stereo camera pair.
	\vspace{-0.1in}
\end{itemize}

\begin{table}[h!]
\begin{center}
\begin{tabular}{|c|c|c|}
\hline
model &  AP @ IoU=0.5 \\
\hline\hline
Image-centric CV Supervised & \textbf{83.6}\\
Image-centric CV Unsupervised & 83.1 \\ \hline
Raycast Stereo w Occupancy &  76.5 \\
Raycast Stereo w/o Occupancy  & 70.5 \\ \hline
Raycast Monocular w/ Occupancy &    56.5\\

\hline
\end{tabular}
\end{center}
\caption{Image centric stereo constraints provide richer object depth information that improves 3D agent detection. The auxiliary task of unsupervised depth estimation achieves similar accuracy to the one supervised with LiDAR depth signal. Occupancy prediction significantly helps the detection task for multi-view stereo setup.
\label{Tab:stereo_vs_psv_mono}}
\end{table}

From our results for the ablation in \tableref{Tab:stereo_vs_psv_mono}, we can draw some observations about the Image-centric models. The first is that the Image-centric Cost volume approach on stereo cameras performs best, beating out the Raycast approach by (+7AP). However this could be due to the difference in auxiliary depth estimation between the two approaches (per-pixel vs occupancy). The next thing we see that that both methods of auxiliary depth estimation for Image-centric CV perform nearly the same, allowing us to provide the additional loss without LiDAR.

If we take a look at the Raycast results, we notice two major points. The first being that the occupancy prediction has a major effect on the stereo model, yielding a +6AP improvement. The second being that having major camera overlap also has large effect on performance. We see a +20AP difference between the monocular and stereo setup on the same data. However we believe that explicit stereo may not be needed for this type of improvement, rather just a strong multiview setup.

\vspace{-0.1in}
\subsubsection*{Training details}
\vspace{-0.1in}

We use the Adam Optimizer for 40 epochs with a base learning rate of 1.25e-4 and use OneCycle learning rate scheduler with batch size 64. We use encoder range of $[-36m, 0m, -1m]$ to $[+36m, +60m, +5m]$ with a voxel resolution of $[0.3m, 0.3m, 0.5m]$. Similar to our panoramic experiments, our stereo models also use the Anchor-Free Detection Head (AFDet)~\cite{afdet} for 3D object detection task and we do not perform any data augmentations during training or test time.
\vspace{-.1in}
\section{Model Analysis}
\vspace{-.1in}

Here we investigate our models from section \ref{panoramic_experiments} in more detail. All metrics were generated on our validation set. For every prediction and ground truth object during evaluation we log statistics such as: object size, distance, number of enclosed LiDAR points and per-camera visibility.

We find that the effects of vision are more prominent at lower LiDAR densities. In \figref{fig:vl_distance}, we see that \textit{Vision + LD} significantly out performs \textit{Vision only} and \textit{LD only}. \textit{Vision + LD} AP performance is even comparable to \textit{HD only} at short ranges (under 35m). At longer ranges, \textit{Vision + LD} sharply degrades, which is consistent with the fact that the panoramic monocular cameras used in this study are fairly short ranged and designed for a wide-surround view. Although the addition of multi-view vision signal to high density LiDAR has less of an effect, the AP performance still outperforms HD only. 
\vspace{-0.1in}
\subsection*{Low Density LiDAR Regime}
\vspace{-0.1in}

\begin{figure}
\centering
\begin{minipage}[t]{.5\textwidth}
  \vspace{-0.1in}
  \centering
  \includegraphics[width=1.0\linewidth]{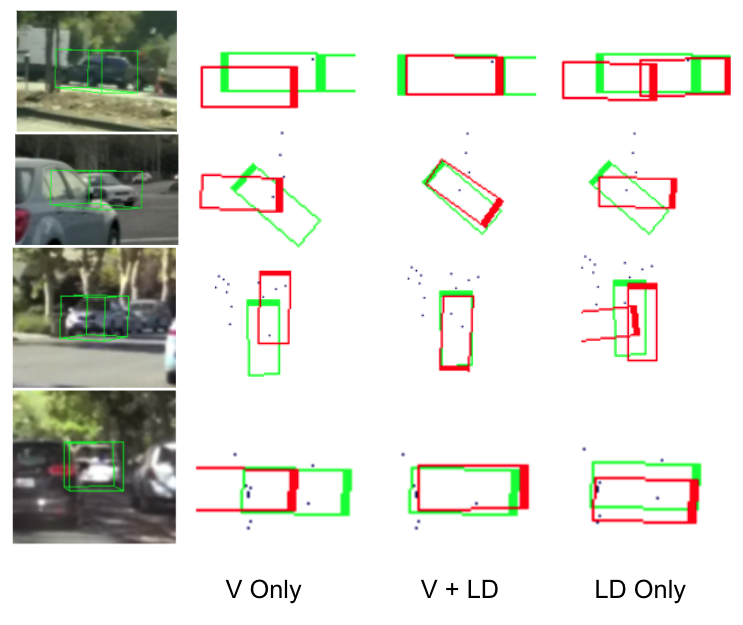}
  \captionof{figure}{Examples where a multi-modal model outperforms single modality models. Groundtruth boxes are green and predicted boxes are red. We see identical detection outputs from a \textit{Vision only} model (left) and a \textit{Vision + LD} model (middle) and a \textit{LD only} model. Isolating objects that have 1 LiDAR point, we see improvements in the fused model where single modality models have failed.}
  \label{fig:single_lidar_point}
\end{minipage}%
\hfill
\begin{minipage}[t]{.45\textwidth}
  \vspace{-0.1in}
  \centering
  \includegraphics[width=0.95\linewidth]{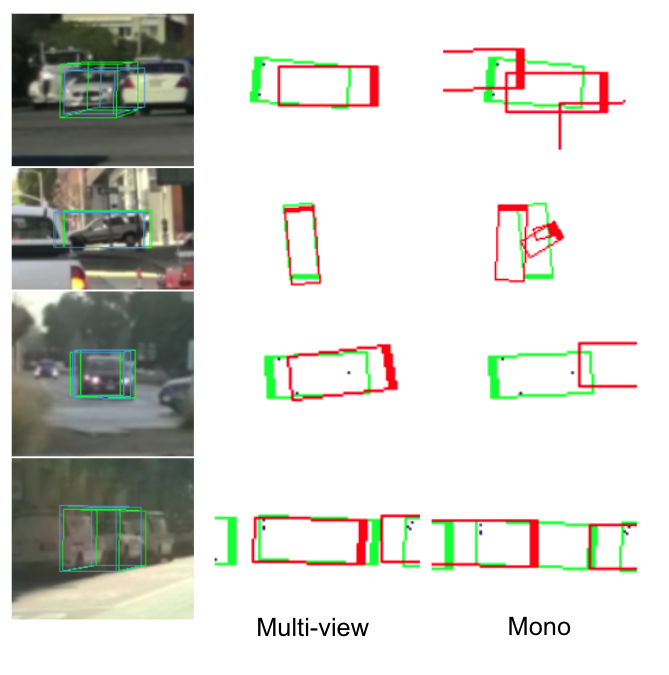}
  \captionof{figure}{Some examples of where multiview improves over monocular baseline; we observe that multiview improves localization error and reduces false positive detections. Grouthtruth boxes are in green and predicted boxes are in red. Here we use the LD dataset but the LiDAR encoder is disabled and is not used in training or evaluation.}
  \label{fig:multiview_mono}
\end{minipage}
\end{figure}

We can take a closer look at the synergy of vision and LiDAR in the low-density (LD) regime. We see in \tableref{lidardataset} that the LD configuration has a median of 1 point per object. We hypothesize that even a few LiDAR points significantly improves depth localization for vision.  This is evident in \figref{fig:vl_distance} where the performance of \textit{Vision only} matches both \textit{LD} and \textit{Vision + LD} at near distances but quickly diverges. As we add more LiDAR points, we achieve longer range detections. We see more evidence of this looking at individual examples where ground truth objects only contain only a single LiDAR point \figref{fig:single_lidar_point}.

\vspace{-0.1in}
\subsection*{Effects of multiview}
\vspace{-0.1in}

In section \ref{multiview_experiments}, we observed that adding an additional view of a scene to Raycasting improves the localization of detections, similar to the effect of adding a few LiDAR points. In \figref{fig:multiview_mono}, we present a subsample of cases where multiview improves the monocular camera baseline. We observe that localization improves in true positive (TP) cases and reduces the number of false positive (FP) boxes around the objects. This is likely attributed to intersection of rays in voxel space, which helps to disambiguate the object location.

\vspace{-.1in}
\section{Conclusion}
\vspace{-.1in}
In this work, we present a modular meta-architecture that allows us to switch between different encoders and modalities to build a generic perception model that enables experimentation with different sensor configurations and deployment across a heterogeneous vehicle fleet. This paradigm allows for a team of model developers that are targeting different AV platforms to easily collaborate on the same model by simply changing parameters. By adding and removing different sensor modalities, we isolate the effects of each sensor. We find that the early fusion of vision and LiDAR is more important at the lower LiDAR densities, and that our low LiDAR density fused model can outperform more expensive sensors in certain cases. Furthermore, we find that having a few LiDAR points on objects significantly increases a vision model's ability to localize objects. We see a similar effect with overlapping cameras in both multiview and stereo setups. Finally, we explore the use of auxiliary depth estimation tasks and find that it improves our vision only models. 

\newpage
\bibliography{egbib}

\end{document}